\documentclass[letterpaper, 10 pt, conference]{ieeeconf}  

\IEEEoverridecommandlockouts                              

\usepackage{graphics} 
\usepackage{graphicx} 
\usepackage{epsfig} 
\usepackage{amsmath, bm} 
\usepackage{amssymb}  
\usepackage{physics}
\usepackage{upgreek}
\usepackage{multicol}
\usepackage{multirow}
\usepackage{lipsum}  
\usepackage{breqn}
\usepackage{comment} 
\usepackage{booktabs}
\usepackage[linesnumbered, ruled]{algorithm2e}
\usepackage{xcolor}

\usepackage{amsfonts}
\usepackage{algorithmic}
\usepackage{textcomp}
\usepackage{xcolor}
\usepackage{soul}

\PassOptionsToPackage{hyphens}{url}

\usepackage[style=ieee,dashed=false]{biblatex}

\DeclareSourcemap{
  \maps{
    \map{
      \pertype{article}
      \step[fieldset=language, null]
      \step[fieldset=url, null]
      \step[fieldset=doi, null]
      \step[fieldset=issn, null]
      \step[fieldset=isbn, null]
      \step[fieldset=note, null]
      \step[fieldset=editor, null]
      \step[fieldset=urldate, null]
      \step[fieldset=file, null]
    }
  }
}
\DeclareSourcemap{
  \maps{
    \map{
      \pertype{inproceedings}
      \step[fieldset=language, null]
      \step[fieldset=url, null]
      \step[fieldset=doi, null]
      \step[fieldset=issn, null]
      \step[fieldset=isbn, null]
      \step[fieldset=note, null]
      \step[fieldset=editor, null]
      \step[fieldset=urldate, null]
      \step[fieldset=file, null]
    }
  }
}
\DeclareSourcemap{
  \maps{
    \map{
      \pertype{incollection}
      \step[fieldset=language, null]
      \step[fieldset=url, null]
      \step[fieldset=doi, null]
      \step[fieldset=issn, null]
      \step[fieldset=isbn, null]
      \step[fieldset=note, null]
      \step[fieldset=editor, null]
      \step[fieldset=urldate, null]
      \step[fieldset=file, null]
    }
  }
}

\title{\LARGE \bf
Quadrupedal Locomotion Control On Inclined Surfaces Using Collocation Method
}

\author{Adarsh Salagame$^{1}$, Maria Gianello$^{1}$, Chenghao Wang$^{1}$, Kaushik Venkatesh$^{1}$,\\ Shreyansh Pitroda$^{1}$, Rohit Rajput$^{1}$, Eric Sihite$^{2}$, Miriam Leeser$^{1}$, and Alireza Ramezani$^{1*}$%
\thanks{$^{1}$ The author is with the Department of Electrical and Computer Engineering, Northeastern University, Boston, MA, USA. (e-mail: a.ramezani@northeastern.edu).}%
\thanks{$^{2}$ The author is with the Department of Aerospace Engineering, California Institute of Technology, Pasadena, USA (e-mail: esihite@caltech.edu)}
\thanks{{*} Corresponding author's e-mail: a.ramezani@northeastern.edu.} 
}

\begin{document}

\maketitle
\thispagestyle{empty}
\pagestyle{empty}

\begin{abstract}
Inspired by Chukars wing-assisted incline running (WAIR), in this work, we employ a high-fidelity model of our Husky Carbon quadrupedal-legged robot to walk over steep slopes of up to 45 degrees. Chukars use the aerodynamic forces generated by their flapping wings to manipulate ground contact forces and traverse steep slopes and even overhangs. By exploiting the thrusters on Husky, we employed a collocation approach to rapidly resolving the joint and thruster actions. Our approach uses a polynomial approximation of the reduced-order dynamics of Husky, called HROM, to quickly and efficiently find optimal control actions that permit high-slope walking without violating friction cone conditions.
\end{abstract}


\section{Wing-Assisted Incline Running (WAIR) Problem}
Robot designs can take many inspirations from nature, where there are many examples of highly resilient and fault-tolerant locomotion strategies to navigate complex terrains by using multi-functional appendages. For instance, Chukars birds perform wing-assisted incline running (WAIR) \cite{10.1242/jeb.001701, dial2003evolution}. These birds showcase impressive dexterity in employing their wings and legs collaboratively to walk-run over steep slopes, resulting in highly plastic locomotion traits which enable them to interact and navigate various environments and expand their habitat range. 

\begin{figure}
    \centering
    \includegraphics[width=1\linewidth]{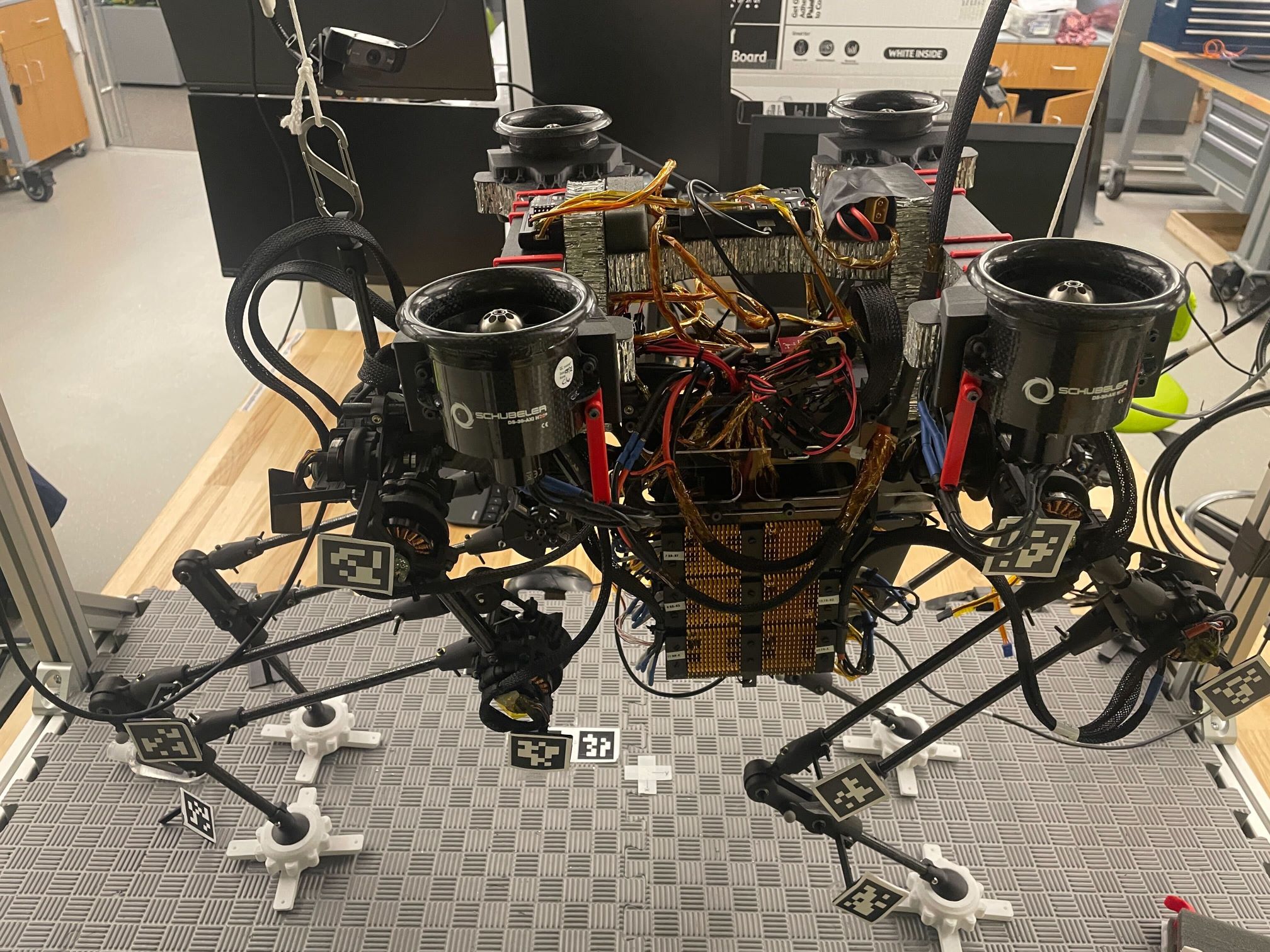}
    \caption{Northeastern University Husky Carbon}
    \label{fig:cover}
\end{figure}

Running over steep slopes can be very challenging since the maximum tangential contact forces allowed can be violated. Chukars recruit their wings to generate external aerodynamic forces in order to manipulate contact forces. The wing-assisted walking can pose rich dynamics and locomotion control problems. In this work, we present our simulated results of the WAIR maneuver using the high-fidelity model of Northeastern's Husky Carbon Robot shown in Fig.~\ref{fig:cover} and a collocation control method with potential applications to real-time control of Husky. 

The robotic biomimicry of animals' multi-modal locomotion can yield mobile robots with unparalleled capabilities. However, among multi-modal animals, those that showcase legged and aerial locomotion pose bigger challenges for robotic biomimicry. Husky's design poses conflicting requirements dictated by ground and aerial locomotion that are not present in, e.g., soft robots which utilize body morphing for mobility \cite{ishida2019morphing, shah2021soft, joyee20223d, kotikian2019untethered}, robots with redundant actuators that combine two separate designs into one single robot (e.g., quad-rotor with wheels \cite{araki2017multi}, or wheeled-legged robots \cite{bjelonic2019keep, grand2004stability, ijspeert2007swimming, riviere2018agile}). 
\nocite{sihite_unsteady_2022, ramezani_towards_2020}

\begin{figure*}
    \centering
    \includegraphics[width=0.7\linewidth]{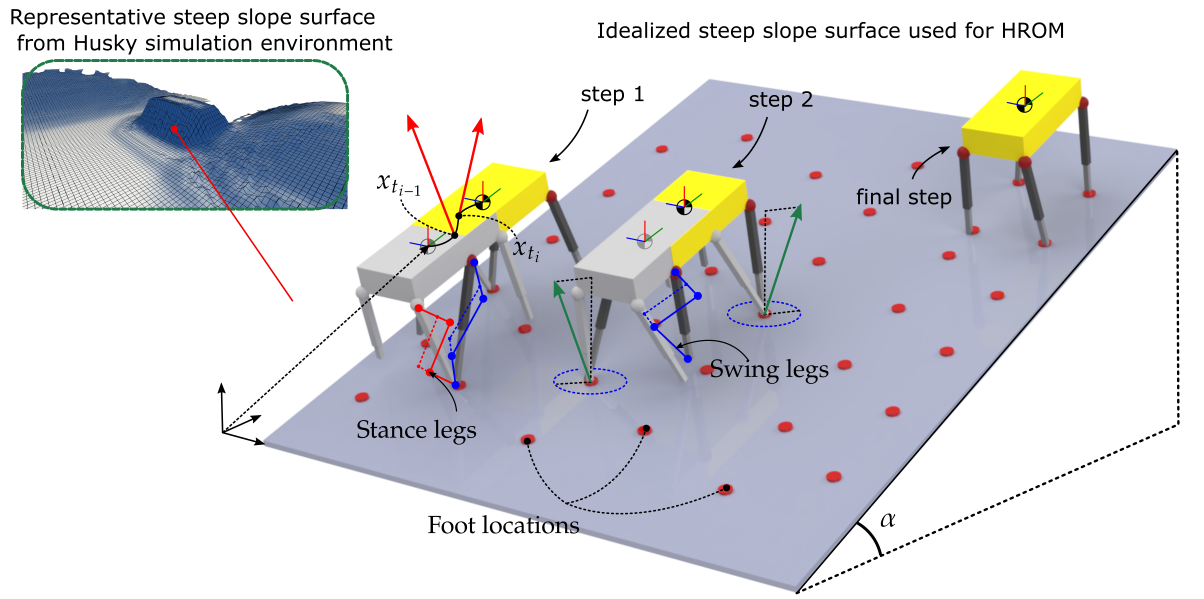}
    \caption{Shows the HROM used to formulate the control design for the WAIR problem.}
    \label{fig:fbd}
\end{figure*}

In this work, we attempt to utilize Husky's thrusters (Fig.~\ref{fig:cover}) to perform WAIR and assist in climbing an inclined surface, as illustrated in Fig.~\ref{fig:fbd}. The addition of thrusters in our robot makes the control design unique compared to previous attempts to climb inclined surfaces with legged or wheeled systems \cite{kim2008development, rigatos2020nonlinear, varghese2017optimal} since in the WAIR maneuver considered for Husky the enforcement of feasible ground contact forces constraints to prevent slippage not only is delivered through dynamic posture control but also involves external thruster force regulations. This problem can be addressed with a plethora of tools from control theory, e.g., optimization-based controllers such as nonlinear model predictive control (MPC) \cite{yue2018efficient}, filter-based techniques such as reference governors \cite{merckaert2018constrained}, or collocation methods \cite{gilday2020vision}. 

\nocite{sihite_unilateral_2021, sihite_orientation_2021, grizzle_progress_nodate, sihite_multi-modal_2023-1, park_finite-state_2013}

This paper's main contribution is the proposition of a novel collocation method based on approximating the dynamics of Husky for rapid control calculations in the WAIR maneuver. We implemented the collocation-based control method to synchronously control joint motions and thrusters forces to traverse steep slopes of 45 degrees without violating the contact conditions. This collocation method will be considered for our real-world tests of the WAIR maneuver using Husky robot.

This paper is organized as follows. The challenges in developing control for WAIR are discussed first, followed by a brief description of Northeastern University's Husky Carbon platform that motivates this work. Then, we describe the simulation model and collocation-based control approach for WAIR. The Result Section presents our closed-loop simulation studies performing WAIR over steep slopes with various inclinations. Finally, we conclude this work with concluding remarks. 

\subsection{Control Challenges of WAIR}

Legged locomotion control on inclined surface can be challenging for known classical problems, e.g., a controller tuned to work for one type of terrain may not work on other terrains and slopes \cite{ma2020quadrupedal}. 
Another classical control challenge is posed by terrain uncertainties. Walking over sloped terrain requires an estimation of the terrain type and the slope itself, after which the robot has to perform an adaptation of the pose of the body and the gait depending on the incline of the slope \cite{gehring2015dynamic, yu2018posture}. 
The absence of vision sensors can make estimation of the slope much harder \cite{zhang2023unknown}. 
The legged locomotion community has tirelessly explored various concepts to democratize adaptive, learning-based, and coupled control approaches to resolve these classical issues facing slope walking \cite{focchi2017high}. 
However, a major research gap remained to solve in sloped terrain legged locomotion is posed by the limitations in legged robots' hardware capabilities, that is, state-of-the-art legged robots can only manipulate contact forces through posture control. This limitation poses major challenges for steep slope terrain locomotion. Instead, in Northeastern's Husky, in addition to a jointed body similar to standard legged robots, four thruster units fixated to the torso allow the direct manipulation of contact forces yielding new control capabilities for steep slope locomotion. In the next section, we briefly describe Husky's hardware design.   

\section{Overview of Northeastern University's \textit{Husky Carbon} Morpho-functional Platform}
\label{sec:husky}

\textit{Husky Carbon} \cite{ramezani2021generative}, shown in Fig.~\ref{fig:cover}, when standing as a quadrupedal robot, is 2.5 ft (0.8 m) tall, is 12 in (0.3 m) wide. The robot was fabricated from reinforced thermoplastic materials through additive manufacturing with a total weight of 9.5 lb (4.3 kg). It hosts on-board power electronics and it operates using an external power supply. The current prototype lacks exteroceptive sensors such as camera and LiDAR. The robot is constructed of two pairs of identical legs in the form of parallelogram mechanisms. Each with three degrees-of-freedom (DOFs), the legs are fixated to Husky's torso by a one-DOF revolute joint with a large range of motion. As a result, the legs can move sideways as well. 

\nocite{lessieur_mechanical_2021, de_oliveira_thruster-assisted_2020}
Sitting atop the robot, attached to the main body is the Propulsion Unit (PU), weighing 7.27 lb (3.3 kg) and bringing the total weight of the robot to 16.77 lb (7.6 kg). The PU consists of four ducted fans, selected for their ability to rapidly change thrust, controlled by a Pixhawk flight controller. It is powered independently by two 6-cell Lithium Polymer batteries and provides an operation time of 4-5 min at full flight level thrusts. Each ducted fan generates a maximum of 2 kg, bringing the total available thrust to about 8 kg, which is just enough to hover fully. The ducted fans are mounted on a very lightweight and strong composite structure made by sandwiching an aluminum honeycomb core between two carbon fiber plates.



\section{WAIR Modeling}
\label{sec:modeling}

\subsection{Husky SimScape Model (Full Dynamics)}

\begin{figure}
    \centering
    \includegraphics[width=0.8\linewidth]{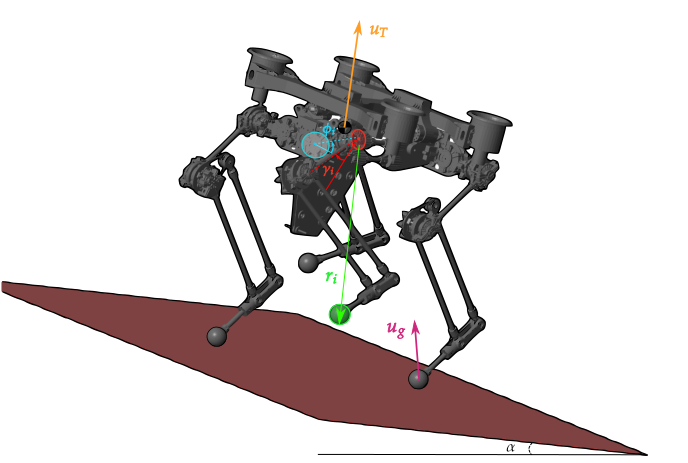}
    \caption{Shows the SimScape model, the model DOFs, and the projection of motions from the SimScape model to HROM.}
    \label{fig:simscape}
\end{figure}

We created a high fidelity model of Husky WAIR model in MATLAB SimScape (see Fig.\ref{fig:simscape}). This model has a total number of 18 degress of freedom (DOF). In this model, we consider a total number of 13 distributed mass elements located at the main body, hips, thighs, and shins. The main body possesses a total mass of 5.0 kg and the diagonal components of mass moment of inertia, $I_{xx}$, $I_{yy}$, $I_{zz}$ ($kg.m^{2}$) $[0.0981867, 0.0844185, 0.164599]$. The hips, thighs, and shins possess body mass of 0.220, 0.060, 0.050 kg. The forward and inverse kinematics of the robot is erected based on the body coordinate frames shown in Fig.~\ref{fig:simscape}. The main body coordinate frame and world frame are linked together using Euler angles.

The sloped surface considered in our analyses is flat with a known slope angle. The ground contact model used in the high-fidelity model is a smooth spring-damper model for normal force, with stiffness $100 N/m$, damping $1 \times 10^{3} N/(m/s)$ and transition region width of $1 \times 10^{-3} m$. And for tangential force, a smooth stick-slip model is used with a coefficient of static friction of $1.8$ and a coefficient of Dynamic Friction of $1.0$.

\subsection{Husky Reduced-Order Model (HROM)}

We use a reduced-order model, called HROM, for the collocation-based control design. In the HROM, each leg is assumed to be massless, that is, all masses are incorporated into the body yielding a 6-DOF model representing the torso's linear and orientation dynamics. Each leg is modeled using two hip angles (frontal and sagittal) and a prismatic joint to describe the leg end position. In HROM, the thruster forces are applied to the body center of mass (COM) and the ground reaction forces (GRF) are applied at the foot end positions. 

Let the superscript $b$ represent a vector defined in the body frame (e.g., $a^b$), and the rotation matrix $R_b\in SO(3)$ represents the rotation of a vector from the body frame to the inertial frame (e.g., $a = R_b a^b$). As such, the foot-end positions in the HROM can be derived using the following kinematics equations:
\begin{equation}
\begin{aligned}
    p_{f_i} &= p_b + R_b l_{h_i}^b + R_b l_{f_i}^b \\
    l_{f_i}^b &= R_{y}\left(\phi_{i}\right) R_{x}\left(\gamma_{i}\right)
    \begin{bmatrix} 
    0, & 0, & -r_{i}
\end{bmatrix}^\top,
\label{eq:foot_pos}
\end{aligned}
\end{equation}
where $p_{f_i}$ and $p_b$ denote the world-frame position of leg-ends and body position, respectively. $l_{h_i}^b$ and $l_{f_i}^b$ denote the body-frame position of hip-COM and foot-hip, respectively. $R_{x}$ and $R_{y}$ denote the rotation matrices around the x- and y-axes. Finally, $\phi_i$ and $\gamma_i$ are the hip frontal and sagittal angles respectively, and $r_i$ is the prismatic joint length.

Let $\omega_b$ be the body angular velocity vector in the body frame and $g$ denote the gravitational acceleration vector. The legs of HROM are massless, so we can ignore all leg states and directly calculate the total kinetic energy $\mathcal{K}= \frac{1}{2}m \dot p_b^{\top} \dot p_b + \frac{1}{2}\omega_b^{\top} J_b \omega_b$ (where $m$ and $J_b$ denote total body mass and mass moment of inertia tensor). The total potential energy of HROM is given by $\mathcal{V}=-m p_b^{\top} g$. Then, the Lagrangian $\mathcal{L}$ of the system can be calculated as $\mathcal{L} = \mathcal{K} - \mathcal{V}$. Hence, the dynamical equations of motion are derived using the Euler-Lagrangian formalism. 

The body orientation is defined using Hamilton's principle of virtual work and the modified Lagrangian for rotation dynamics in SO(3) to avoid using Euler rotations which can become singular during the simulation. The equations of motion for HROM are given by
\begin{equation}
\begin{gathered}
    \textstyle \frac{d}{d t} \left( \frac{ \partial \mathcal{L}}{\partial \dot p_b } \right ) - \frac{\partial \mathcal{L}}{\partial p_b} = f_{gen}, \qquad
    \dot{R}_b = R_b\,[\omega_b]_\times \\
    \textstyle \frac{d}{dt}\left( \frac{\partial \mathcal{L}}{\partial \omega_b}  \right) + 
    \omega_b \times \frac{\partial \mathcal{L}}{\partial \omega_b} + 
    \sum_{j=1}^{3} r_{b_j} \times \frac{\partial \mathcal{L}}{\partial r_{b_j}} = \tau_{gen},
\end{gathered}
\label{eq:euler-lagrangian}
\end{equation}
where $f_{gen}$ and $\tau_{gen}$ are the the generalized forces and moments (from GRF and thrusters), $[\,\cdot\,]_\times$ is the skew operator, and $R_b^\top = [r_{b_1}, r_{b_2}, r_{b_3}]$ (i.e., $r_{b_j}$ are the columns of $R_b$). The HROM model can then be solved from into the following standard form:
\begin{equation}
M\,
\begin{bmatrix}
\ddot p_b \\ \dot \omega_b
\end{bmatrix} 
+ H = \textstyle \sum_{i=1}^{4} B_{g_i}\, u_{g_i} + B_T\, u_T,
\label{eq:eom_dynamics}
\end{equation}
where $M$ is the mass-inertia matrix, $H$ contains the Coriolis matrix and gravity vector, $B_{g_i} u_{g_i}$ represent the generalized force due to the GRF $u_{g_i}$ acting on the i-th foot where $B_{g_i} = \Big(\frac{\partial \dot p_{f_i}}{\partial v}\Big)^\top$ and $v = [\dot p_b^\top, \omega_b^\top]^\top$. In Eq.~\ref{eq:eom_dynamics}, $u_T$ denotes the thruster action and $B_T= \Big(\frac{\partial \dot p_b}{\partial v}\Big)^\top$.

Note that the matrix $M$, $H$, $B_{g_i}$, and $B_T$ are functions of the robot's posture, that is, the leg joint variables $\phi_i$, $\gamma_i$, and $r_i$. The way we control posture is achieved by finding the proper accelerations for the desired body motion which is addressed by the collocation-based method as discussed later. The joint states and the inputs are defined as follows
\begin{equation}
    q_L = [\dots,\phi_i, \gamma_i, r_i,\dots]^\top, \qquad
    \ddot q_L = u_L
\end{equation}
Combining both the body dynamics given by Eq.~\ref{eq:eom_dynamics} and massless-leg states given above form the HROM state vector which is given by:
\begin{equation}
    x = [q_L^\top, \dot q_L^\top, r_b^\top, p_b^\top, \omega_b^\top, \dot p_b^\top]^\top,
\end{equation}
where the $r_b$ denotes the columns of the rotation matrix $R_b$. The state-space model of HROM can be defined as 
\begin{equation}
    \dot x = f_{_{ROM}}(x, u_L, u_g, u_T)
    \label{eq:hrom-state-space}
\end{equation}
which will be utilized for WAIR control of the high fidelity model in SimScape.

\subsection{HROM Ground Contact Model}

The ground model used in the HROM is fined tuned to match the model in the SimScape model. This section, briefly explains the HROM ground contact dynamics. The ground model used in HROM is given by 
\begin{equation}
\begin{aligned}
    u_{g_i} &= \begin{cases} \, 0 ~~  \mbox{if } p_{f_{i,z}} > 0  \\
     [u_{g_{i,x}},\, u_{g_{i,y}},\, u_{g_{i,z}}]^\top ~~ \mbox{else} \end{cases} \\
    u_{g_{i,z}} &= -k_1 p_{f_{i,z}} - k_{2} \dot p_{f_{i,z}} \\
    u_{g_{i,j}} &= - s_{j} u_{g_{i,z}} \, \mathrm{sgn}(\dot p_{f_{i,j}}) - \mu_v \dot p_{f_{i,j}} ~~  \mbox{if} ~~j=x, y\\
    s_{j} &= \Big(\mu_c - (\mu_c - \mu_s) \mathrm{exp} \left(\frac{-|\dot p_{f_{i,j}}|^2}{v_s^2}  \right) \Big)
\end{aligned}
\end{equation}
\noindent where $p_{f_{i,j}},~~j=x,y,z$ are the $x-y-z$ positions of the contact point; $u_{g_{i,j}},~~i=x,y,z$ are the $x-y-z$ components of the ground reaction force assuming a point contact takes place between the robot and the ground substrate; $k_{1}$ and $k_{2}$ are the spring and damping coefficients of the compliant surface model; $\mu_c$, $\mu_s$, and $\mu_v$ are the Coulomb, static, and viscous friction coefficients; and, $v_s > 0$ is the Stribeck velocity.

\section{Control}

Consider $N$ time intervals during the WAIR maneuver 
\begin{equation}
0=t_1<t_2<\ldots<t_N=t_f
    \label{eq:?}
\end{equation}
\noindent where $t_k$ denotes discrete times. First, we discretize the continuous model given by Eq.~\ref{eq:hrom-state-space} using an explicit Euler integration scheme as follows
\begin{equation}
\begin{aligned}
    x_{k+1} &= x_{k} + \Delta t f_{_{ROM}} (x_{k},u_{L,k}, u_{g,k}, u_{T,k}) , \\
    &\quad k=1, \ldots, N, \quad 0 \leq t_k \leq t_f
\end{aligned} 
\label{eq:discrete_model}
\end{equation}   
where $\Delta t$ is the integration time step, $t_k$ is the discrete time at k-th discrete step, $x_k = x(t_k)$ denotes the state vector at $t_k$, and $u_{i,k} = u_{i,k}(t_k)$ (where $i=L,~g,~T$) is the input. Let $x_{r,k}$ be the state reference and $e_k = x_{r,k} - x_{k}$ be the tracking error for the states of HROM. Then, we consider the following cost function 
\begin{equation}
\begin{aligned}
    J = \sum_{k=1}^{N} \left( e_k^\top \,Q\, e_k + u_{k-1}^\top \,R\, u_{k-1} \right)
\end{aligned} 
\label{eq:cost}
\end{equation}   
where $N$ is ..., $Q$ and $R$ denote the diagonal weighting matrices used to penalize the tracking performance and control efforts, respectively. Note that from here onward as shown in Eq.~\ref{eq:cost}, all inputs, including the GRF, joint accelerations, and thruster forces, are stacked in the input vector $u$.

Our objective is to find $u_k$ based on cubic collocation at Lobatto points such that $J$ is minimized during the WAIR maneuver. We consider 2N boundary conditions given by
\begin{equation}
    \begin{aligned}
        r_i\left(x(0), x\left(t_f\right), t_f\right)=&0
    \end{aligned}
\end{equation}
\noindent to enforce continuity of the state vector evolution. We consider N inequality constraints given by
\begin{equation}
    g_i(x(t), u(t), t) \geq 0
\end{equation}
\noindent to limit the input $u_k$ in each discrete time period. We stack all of the discrete states $x_k$ and inputs $u_k$ from the HROM model given by Eq.~\ref{eq:discrete_model} in the vectors $X = \left[x^\top_1, \ldots, x^\top_k\right]^\top$ and $U = \left[u^\top_1, \ldots, u^\top_k\right]^\top$. In this work, we resolve the optimal solutions for the WAIR maneuver for fixed $t_f$. However, it is possible $t_f$, as the decision parameter in the optimization problem, that is, we add final discrete time $t_f$ as the last entry of the decision parameter vector $Y$,
\begin{equation}
Y=\left(X,U, t_f\right) \in \mathbb{R}^{2N+1}
    \label{?}
\end{equation}
We find $X$ and $U$ as the decision parameters of the optimization problem using MATLAB's nonlinear optimization toolbox. To resolve the optimization problem rapidly, we employ an interpolation approach to approximate $x_k$ and $u_k$. We take input to be as the linear interpolation function between $u(t_k)$ and $u(t_{k+1})$ for $t_k \leq t<t_{k+1}$, that is, $u_{int}$ is given by
\begin{equation}
u_{int}(t)=u\left(t_k\right)+\frac{t-t_k}{\Delta t}\Big(u\left(t_{k+1}\right)-u\left(t_k\right)\Big)
    \label{}
\end{equation}
\noindent In addition, we interpolate the HROM's states $x(t_k)$ and $x(t_{k+1})$ too. This way, the speed of control computations increases considerably. We take a nonlinear cubic interpolation which is continuously differentiable with 
\begin{equation}
    \dot{x}_{\mathrm{int}}(s)=f_{_{ROM}}(x(s), u(s), s)
\end{equation}
at $s=t_k$ and $s=t_{k+1}$. To do this, we write the following system of equations:
\begin{equation}
    \begin{aligned}
x_{int}(t) &=\sum_{k=0}^3 c_k^j\left(\frac{t-t_j}{h_j}\right)^k, \quad t_j \leq t<t_{j+1}, \\
c_0^j &=x\left(t_j\right), \\
c_1^j &=h_j f_j, \\
c_2^j &=-3 x\left(t_j\right)-2 h_j f_j+3 x\left(t_{j+1}\right)-h_j f_{j+1}, \\
c_3^j &=2 x\left(t_j\right)+h_j f_j-2 x\left(t_{j+1}\right)+h_j f_{j+1}, \\
\text { \textbf{where} } f_j &=f_{_{ROM}}\left(x\left(t_j\right), u\left(t_j\right), t_j\right),\\
h_j&=t_{j+1}-t_j .
\end{aligned}
\label{eq:cubic-lobatto}
\end{equation}

\begin{figure*}
    \centering
    \includegraphics[width=0.65\linewidth]{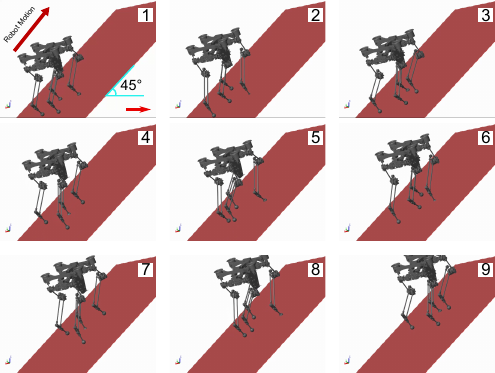}
    \caption{Simulated snapshots of the WAIR maneuver on a 45-deg slope using Husky's high-fidelity model in MATLAB SimScape.}
    \label{fig:wair_sim}
\end{figure*}

The interpolation function used for $x_{int}$ must satisfy the derivatives at the discrete points and at the middle of sample times $t_{c,i}$. 

By inspecting Eq.~\ref{eq:cubic-lobatto}, it can be seen that the derivative terms at the boundaries $t_{i}$ and $t_{i+1}$ are satisfied. Therefore, the only remaining constraints in the nonlinear programming constitute the collocation constraints at the middle of $t_i-t_{i+1}$ time interval, the inequality constraints at $t_i$, and the constraints at $t_1$ and $t_f$. This property of the interpolation method given by Eq.~\ref{eq:cubic-lobatto} reduces the total number of equations that must be resolved yielding a considerable speed in control computations. Hence, the remaining constraints in the nonlinear programming are given by:
\begin{equation}
\begin{aligned}
f_{_{ROM}}\Big(x_{int}\left(t_{c,i}\right), u_{int}\left(t_{c,i}\right)\Big)-\dot x_{int}\left(t_{c,i}\right)&=0\\
g\Big(x_{int}\left(t_i\right), u_{int}\left(t_i\right), t_i\Big) &\geq 0\\
r\Big(x_{int}\left(t_1\right), x_{int}\left(t_N\right), t_N\Big)&=0
\end{aligned}
    \label{eq:fdc-controller}
\end{equation}
\noindent Given that the computational structure is spatially discrete with large costs associated with its curse of dimensionality, this collocation scheme results in a smaller number of parameters for interpolation polynomials which enhance the computation performance. We resolve this optimization problem using MATLAB fmincon function.  

\section{Results}
\label{sec:results}
\begin{figure}
    \centering
    \includegraphics[width=0.8\linewidth]{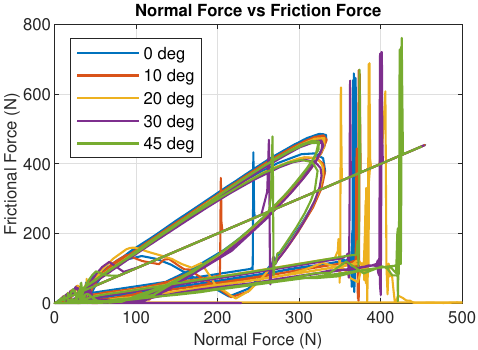}
    \caption{Illustrates the normal force $u_{g_{i,z}}$ versus tangential $\sqrt{u^2_{g_{i,x}}+u^2_{g_{i,y}}}$ at the stance leg-ends during the WAIR maneuver.}
    \label{fig:grf2}
\end{figure}
\begin{figure}
    \centering
    \includegraphics[width=0.8\linewidth]{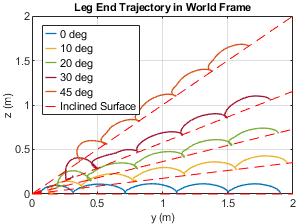}
    \caption{Show the swing leg-end trajectory during slope climbing. The dashed red line denotes the slope surface.}
    \label{fig:leg_end_pos}
\end{figure}
\begin{figure}
    \centering
    \includegraphics[width=0.8\linewidth]{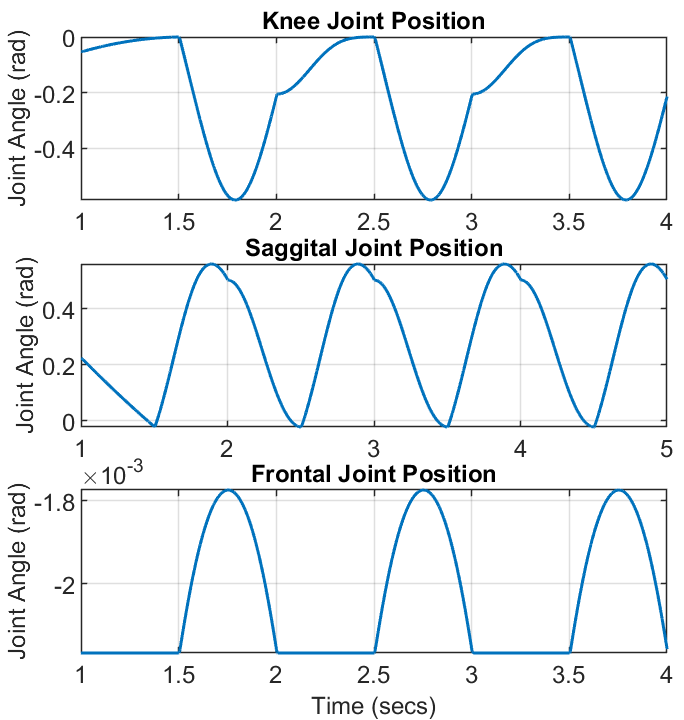}
    \caption{Show the joint angles for the front leg during slope climbing.}
    \label{fig:joint_position}
\end{figure}
\begin{figure}
    \centering
    \includegraphics[width=0.8\linewidth]{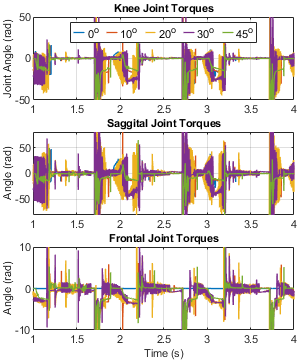}
    \caption{Show the joint torques for the front legs during slope climbing.}
    \label{fig:joint_torque}
\end{figure}

We simulated the closed-loop WAIR problem. Figure~\ref{fig:wair_sim} shows the snapshots of the high-fidelity SimScape model. In this simulation, Husky walks up a 45-deg incline. Other WAIR simulations include walking over 0 deg, 10 deg, 20 deg, and 30 deg inclinations. The gait remained fixed in these simulations, as shown in Figure \ref{fig:joint_position}. Figure~\ref{fig:leg_end_pos} shows the swing leg-end positions for different inclinations. The manipulated ground contact forces for different inclination scenario satisfies friction cone conditions shown in Fig.~\ref{fig:grf2}. The joint torques, shown in Fig.~\ref{fig:joint_torque}, show higher peaks at lower inclinations as the legs carry a larger percentage of the total body weight and thruster actions contribute less. However, in higher inclinations, thruster contributions become larger.

\section{Concluding Remarks}
\label{sec:conlusion}
Inspired by Chukars wing-assisted incline running (WAIR), in this work, we employed a high-fidelity model of our Husky Carbon quadrupedal-legged robot to study the control design for steep slope walking. Chukars use the aerodynamic forces generated by their flapping wings to manipulate ground contact forces and traverse steep inclines and even overhangs. By exploiting the unique design of Husky, we employed a collocation approach to resolving the joint and thruster actions rapidly. In our approach, we used a polynomial approximation of the reduced-order dynamics of Husky, called HROM, to quickly and efficiently find optimal control actions that permit high-slope walking without violating friction cone conditions. Future works will focus on implementing the WAIR controller on Husky Carbon.




\printbibliography

\end{document}